\documentclass[11pt,a4paper]{article}
\usepackage[hyperref]{acl2020}
\usepackage{times}
\usepackage{latexsym}

\usepackage{array}

\usepackage{microtype}

\aclfinalcopy % Uncomment this line for the final submission
\title{Unsupervised Technical Domain Terms Extraction using Term Extractor}

\author{Suman Dowlagar \\
  LTRC\\
  IIIT-Hyderabad\\
  \texttt{suman.dowlagar@} \\
  \texttt{research.iiit.ac.in} \\\And
  Radhika Mamidi \\
  LTRC\\
  IIIT-Hyderabad\\
  \texttt{radhika.mamidi@} \\
  \texttt{iiit.ac.in} \\}

\date{}

\begin{document}
\maketitle
\begin{abstract}
Terminology extraction, also known as term extraction, is a subtask of information extraction. The goal of terminology extraction is to extract relevant words or phrases from a given corpus automatically. This paper focuses on the unsupervised automated domain term extraction method that considers chunking, preprocessing, and ranking domain-specific terms using relevance and cohesion functions for ICON 2020 shared task 2: TermTraction.
\end{abstract}

\section{Introduction}

The aim of Automatic Term Extraction (ATE) is to extract terms such as words, phrases, or multi-word expressions from the given corpus. ATE is widely used in many NLP tasks, such as machine translation, summarization, clustering the documents, and information retrieval. 

Unsupervised algorithms for domain term extraction are not labeled and trained on the corpus and do not have any pre-defined rules or dictionaries. They often use statistical information from the text. Most of these algorithms use stop word lists and can be applied to any text datasets.  The standard unsupervised automated term extraction pipeline consists of 
\begin{itemize}
\item Simple Rules: using chunking or POS tagging to extract Noun phrases for multi-word extraction.
\item Naive counting: that counts how many terms each word occurs in the corpus.
\item Preprocessing: Removing punctuation and common words such as stop words from the text.
\item Candidate generation and scoring: using statistical measures and ranking algorithms to generate the possible set of domain terms
\item Final set: Arrange the ranked terms in descending order based on the scores and take the top N keywords as the output.
\end{itemize}

Currently, there are many methods for automatic term recognition. \citet{evans1995clarit} used TF-IDF measure for term extraction.\citet{navigli2002semantic} used domain consensus which is designed to recognize the terms uniformly distributed over the whole corpus. The most popular method C-value \cite{frantzi2000automatic} is also a statistical measure that extracts a term based on the term's frequency, length of the term, and the set of the candidates that
enclose the term such that the term is in their substring.  \citet{bordea2013domain} proposed the method called Basic, which is a modification of the C-value for recognizing
terms of average specificity. The successor of C-value statistic called the NC value \cite{frantzi2000automatic} considered scored the term based on the condition if it exists in a group of common words or if it contains nouns, verbs, or adjectives that immediately precede or follow the term. The methods proposed by  \citet{ahmad1999university,kozakov2004glossary,sclano2007termextractor} are based on extracting the terms of a text by considering the frequency of occurrence of terms in the general domain.

A detailed survey of the existing automated term extraction algorithms and their evaluation are presented in papers by  \citet{astrakhantsev2015methods,vsajatovic2019evaluating}

In this paper, we used the term extractor algorithm \cite{sclano2007termextractor} present in the pyate\footnote{https://pypi.org/project/pyate/} library for domain term extraction. The term extractor algorithm is developed initially for ontology extraction from large corpora. It uses domain pertinence/relevance, domain consensus, and lexical cohesion for extracting terms. A detailed description of the modules is given in the next section.

The paper is organized as follows. Section 2 gives a detailed description of the term extraction algorithm used. Section 3 gives information about the datasets used and results. Section 4 concludes the paper. 

\section{Our Approach}

In this section, we describe in detail the methods used in the term extractor algorithm.

Initially, TermExtractor performs chunking and proper name recognition and then extracts structures based on linguistic rules and patterns, including stop words, detection of misspellings, and acronyms. The extraction algorithm uses Domain Pertinence, Domain Cohesion, and Lexical Cohesion to decide if a term is considered a domain term. 

Domain Pertinence, or Domain Relevance (DR), requires a contrastive corpus and compares a candidate's occurrence in the documents belonging to the target domain to its occurrence in other domains, but the measure only depends on the contrastive domain where the candidate has the highest frequency. The Domain Pertinence is based on a simple formula,

\begin{equation}
 DR_{D_i}(t) = \frac{tf_i}{max_j (tf_j )}
\end{equation}

Where $tf_i$ is the frequency of the candidate term in the input domain-specific document collection and $max_j (tf_j )$ is the general corpus domain, where the candidate has the highest frequency, and $D_i$ is the domain in consideration.

Domain Consensus (DC) assumes that several documents represent a domain. It measures the extent to which the candidate is evenly distributed on these documents by considering normalized term frequencies $(\phi)$, 

\begin{equation}
DC_{D_i}(t) = \sum_{d_k \epsilon D_i}\phi_klog\phi_k
\end{equation}

Here, we assume $k$ distinct documents for the domain $D_i$.

Lexical cohesion involves the choice of vocabulary. It is concerned with the relationship that exists between lexical items in a text, such as words and phrases. It compares the in-term distribution of words that make up a term with their out-of-term distribution.

\begin{equation}
 LC_{D_i}(t) = \frac{n*tf_i*logtf_i}{\sum_{j}tf_{w_j}i}
\end{equation}

Where $n$ is the number of documents in which the term $t$ occurs.

The final weight of a term is computed as a weighted average of the three filters above,

\begin{equation}
 score(t,D_i) = \alpha*DR+\beta*DC+\gamma*LC
\end{equation}

where $\alpha$, $\beta$, $\gamma$ are the weights, and they are equal to $1/3$

\section{Experiments}

This section describes the dataset used for domain terms extraction, implementation of the above approach, followed by results, and error analysis.

\subsection{Dataset}

We used the dataset provided by the organizers of TermTraction ICON-2020. The task is to extract domain terms from the given English documents from the four technical domains like Computer Science, Physics, Life Science, Law. The data statistics of the documents in the respective domains are shown in the table \ref{tab:data}.

\begin{table}[]
\begin{tabular}{|l|l|l|}
\hline
\textbf{Domain}           & \textbf{\#Train docs} & \textbf{\#Test docs} \\
\hline
\textit{Bio-Chemistry}    & 229                      & 10                       \\
\textit{Communication}    & 127                      & 10                     \\
\textit{Computer-Science} & 201                      & 8                      \\
\textit{Law}              & 70                      & 16                     \\
\hline
\end{tabular}
\caption{Data statistics}
\label{tab:data}
\end{table}

\begin{table*}[!t]
\centering
\begin{tabular}{|p{1cm}p{0.8cm}p{0.8cm}|p{1.2cm}p{0.8cm}p{0.8cm}|p{1.1cm}p{0.8cm}p{0.8cm}|p{0.8cm}p{0.8cm}p{0.8cm}|}
\hline
\multicolumn{3}{|c|}{\textbf{Biochemistry}}        & \multicolumn{3}{c|}{\textbf{Communication}}       & \multicolumn{3}{c|}{\textbf{Computer Science}}    & \multicolumn{3}{c|}{\textbf{Law}}                \\
\hline
\textbf{Data}  & \textbf{run 1} & \textbf{run 2} & \textbf{Data}  & \textbf{run 1} & \textbf{run 2} & \textbf{Data}  & \textbf{run 1} & \textbf{run 2} & \textbf{Data} & \textbf{run 1} & \textbf{run 2} \\
\hline
\textbf{M12S1} & 0.247          & 0.222          & \textbf{M2-1}  & 0.109          & 0.086          & \textbf{KL2}   & 0.220          & 0.225          & \textbf{A01}  & 0.079          & 0.077          \\
\textbf{M15S2} & 0.208          & 0.195          & \textbf{M2-2}  & 0.102          & 0.104          & \textbf{KL4}   & 0.241          & 0.246          & \textbf{A02}  & 0.099          & 0.066          \\
\textbf{M16S2} & 0.224          & 0.207          & \textbf{M2-3}  & 0.094          & 0.074          & \textbf{KL8}   & 0.138          & 0.146          & \textbf{A03}  & 0.144          & 0.126          \\
\textbf{M23S3} & 0.266          & 0.233          & \textbf{M3-1}  & 0.240          & 0.236          & \textbf{W12}   & 0.143          & 0.122          & \textbf{FA1}  & 0.104          & 0.116          \\
\textbf{M26S2} & 0.096          & 0.081          & \textbf{M3-2}  & 0.159          & 0.148          & \textbf{W1332} & 0.216          & 0.195          & \textbf{FA2}  & 0.077          & 0.067          \\
\textbf{T18}   & 0.463          & 0.427          & \textbf{M3-3}  & 0.140          & 0.132          & \textbf{W13}   & 0.108          & 0.089          & \textbf{FC1}  & 0.082          & 0.073          \\
\textbf{T25}   & 0.310          & 0.282          & \textbf{RM16}  & 0.101          & 0.088          & \textbf{W1436} & 0.181          & 0.165          & \textbf{FC2}  & 0.032          & 0.021          \\
\textbf{T39}   & 0.265          & 0.247          & \textbf{RM17}  & 0.067          & 0.065          & \textbf{W921}  & 0.221          & 0.188          & \textbf{FC3}  & 0.016          & 0.014          \\
\textbf{T4}    & 0.271          & 0.234          & \textbf{RM18}  & 0.098          & 0.115          &                &                &                & \textbf{FR1}  & 0.149          & 0.113          \\
\textbf{T9}    & 0.323          & 0.315          & \textbf{SW1AW} & 0.120          & 0.113          &                &                &                & \textbf{FR2}  & 0.144          & 0.112          \\
               &                &                &                &                &                &                &                &                & \textbf{FR3}  & 0.073          & 0.062          \\
               &                &                &                &                &                &                &                &                & \textbf{G3}   & 0.103          & 0.098          \\
               &                &                &                &                &                &                &                &                & \textbf{G4}   & 0.056          & 0.052          \\
               &                &                &                &                &                &                &                &                & \textbf{R1}   & 0.022          & 0.055          \\
               &                &                &                &                &                &                &                &                & \textbf{R2}   & 0.033          & 0.026          \\
               &                &                &                &                &                &                &                &                & \textbf{R3}   & 0.044          & 0.048      \\
\hline
\end{tabular}
\caption{Term Extraction macro-F1 score.}
\label{tab:res}
\end{table*}

\begin{table*}[!t]
\centering
\begin{tabular}{|m{11cm}|l|}
\hline
\textbf{Template Sentence}                                                                                                                                                                                                                                                                                                                                                                                                                                                                                                                                                                                                            & \textbf{Domain terms identified}                                                                                                                                                                                                                                                                            \\
\hline
We are not going to that , remove it completely, but nevertheless this is an indication that , \textbf{NO plus} is going to be a \textbf{poorer donor} , compared to \textbf{carbon monoxide} . So , this drastic reduction in the \textbf{stretching frequency} can only happen if you have , a large population of the \textbf{anti - bonding orbitals} of \textbf{NO plus} . And it has got a structure , which is very similar , a structure which is very similar to the structure of \textbf{nickel tetra carbonyl} . You will see that , while \textbf{carbon monoxide} is ionized with 15 \textbf{electron volts} , if you supply 15 \textbf{electron volts} , \textbf{carbon monoxide} can be \textbf{oxidized} or \textbf{ionized} . & \begin{tabular}[c]{@{}l@{}}large population \\ similar \\ ionized \\ carbonyl \\ frequency \\ poorer donor \\ anti - bonding orbitals \\ indication \\ carbon monoxide \\ electron volts \\ nickel \\ plus \\ drastic reduction \\ structure\end{tabular}\\
\hline
\end{tabular}
\caption{Error analysis on the template sentence}
\label{tab:template_data}
\end{table*}

\subsection{Implementation}

We used Pyate (python automated term extraction library) that contains the term extractor method and is trained on the general corpus. With the help of the term extraction method, we extracted the relevant terms from the given corpus.

We have submitted two runs, one run (run 1) is the term extractor function itself, and the other run (run 2) is term extractor combined with NP chunks of phrase length > 2 obtained from NLTK ConsecutiveNPChunkTagger\footnote{ConsecutiveNPChunkTagger
}.

\subsection{Results and Error Analysis}

We evaluated the performance of the method using average precision. The results are tabulated in Table \ref{tab:res}.

For the template sentence given in Table \ref{tab:template_data}, our algorithm  failed to recognize the domain terms \textbf{NO plus} and \textbf{nickel tetra carbonyl}. It considered \textbf{NO} as the stop word (no or negation) and discarded it while preprocessing. The algorithm also misunderstood words like ``similar'' as domain terms and failed to identify \textbf{nickel tetra carbonyl} as a domain term. It indicates that further study is necessary, which considers the candidate terms' capitalization and uses better methods that support the more reliable form of compound words or multi-word expressions.

\section{Conclusion}
For domain term extraction from technical domains like Bio-Chemistry, Law, Computer-Science, and communication, We used the term extractor method from pyate library for obtaining technical terms. The term extractor method uses keywords from the general corpora, and it considers Domain Pertinence, Domain Cohesion, and Lexical Cohesion methods for extracting domain terms in the given corpus. 

As mentioned above, it did not give preference to capitalized terms and did not consider some compound words. So we have to work towards better methods that consider capitalization, better formation of compound words for the more reliable performance of the automated domain term extractor.

\bibliography{anthology,acl2020}
\bibliographystyle{acl_natbib}

\appendix

\end{document}